\documentclass[runningheads]{llncs}
\usepackage[T1]{fontenc}
\usepackage{graphicx}
\usepackage{subfig}
\usepackage{subcaption}
\usepackage{svg}
\usepackage[utf8]{inputenc} 
\usepackage[T1]{fontenc}    
\usepackage{hyperref}       
\usepackage{url}            
\usepackage{booktabs}       
\usepackage{amsfonts}       
\usepackage{nicefrac}       
\usepackage{microtype}      
\usepackage{xcolor}         
\usepackage{pifont}

\usepackage{xspace}
\newcommand*{\eg}{e.g.\@\xspace}

\newcommand*{\etc}{%
    \@ifnextchar{.}%
        {etc}%
        {etc.\@\xspace}%
}

\newcommand{\cmark}{\ding{51}}%
\newcommand{\xmark}{\ding{55}}%

\newcommand{\cG}{\mathcal{G}}

\newcommand{\cA}{\mathcal{A}}
\newcommand{\ba}{\mathbf{a}}

\newcommand{\cC}{\mathcal{C}}

\newcommand{\cP}{\mathcal{P}}

\newcommand{\bc}{\mathbf{c}}

\newcommand{\bg}{\mathbf{g}}

\usepackage{afterpage}

\newcommand{\bp}{\mathbf{p}}

\newcommand{\be}{\mathbf{e}}

\begin{document}

\title{Finding Outliers in a Haystack: \newline Anomaly Detection for Large Pointcloud Scenes}
\titlerunning{Finding Outliers in a Haystack}
%
\author{Ryan Faulkner, Luke Haub, Simon Ratcliffe, Tat-Jun Chin}
%
\institute{University of Adelaide, Australia \and
Maptek, Australia}
%
\maketitle              
\begin{abstract}
LiDAR scanning in outdoor scenes acquires accurate distance measurements over wide areas, producing large-scale point clouds. Application examples for this data include robotics, automotive vehicles, and land surveillance. During such applications, outlier objects from outside the training data will inevitably appear. Our research contributes a novel approach to open-set segmentation, leveraging the learnings of object defect-detection research. We also draw on the Mamba architecture's strong performance in utilising long-range dependencies and scalability to large data. Combining both, we create a reconstruction based approach for the task of outdoor scene open-set segmentation. We show that our approach improves performance not only when applied to our our own open-set segmentation method, but also when applied to existing methods. Furthermore we contribute a Mamba based architecture which is competitive with existing voxel-convolution based methods on challenging, large-scale pointclouds.

\keywords{3D Vision  \and Scene Understanding \and Anomaly Detection \and Open Set Segmentation}
\end{abstract}

\section{Introduction}

Understanding large pointcloud scenes is an important goal for a variety of applications, from self-driving vehicles or drones, to general land surveillance. Accurate three-dimensional data can be obtained from Light Detection and Ranging (LiDAR) scanners, offering geometric information and precision which a two dimensional camera cannot. However between the structure of the data, and the potential scale, with millions of points and objects over a hundred meters apart, processing these LiDAR scans can be difficult. Furthermore, when the context of a practical application is as uncontrolled and diverse as an outdoor scene, there will inevitably be anomalous objects which were not in the training data. Identifying these unknown anomalies when they appear is essential to the reliable application of theoretical architectures and methods to real-world applications.

Existing ``Anomaly Detection'' research can be grouped into two distinct categories. Either detecting defects and flaws when examining a single object at a time\cite{MambaAD,ALMRR}, or identifying unknown objects which were not present in the training data, in a scene context\cite{GANReconstructFromSegmentation,doss,real}. We develop a novel approach to open-set segmentation, developed by approaching the problem of anomalous object detection from both of these perspectives. 
\newline

Instead of defining the problem as either "Identifying \textit{outliers} within a consistent object" or "Identifying \textit{unknown} objects not present in training" we combine the two, and define the problem as "Identifying \textit{unknown outliers} in a scene". Our paper puts forward that by training a reconstruction step using only known classes in a scene, we have developed an architecture which achieves superior performance at detecting unknown objects (anomalies).

In this paper we provide:

\begin{itemize}
    \item One mamba-based architecture for reconstructing a scene's ``default context'' trained only on known objects. 
    \item A second architecture for the task of anomaly detection. Our mamba backbone contrasts to existing voxel convolution based methods. 
    \item Analysis and ablation testing provides the first exploration of how Mamba, which boasts high efficiency and ability to utilise long-range dependencies in data, applies to the field of anomaly detection for large-scale point cloud data.
    \item We propose a novel method for creating synthetic anomalies to support training.
\end{itemize}


\section{Related Work}

\subsection{Open-Set Segmentation for Point Cloud Scenes}
There are some methods which fall outside of machine learning/AI which identify anomalous objects. These include representing the data as a graph, within which ``outliers'' can be identified \cite{graphSignal}. Alternatively, using predefined matrices and equations (instead of learned ones) to extract object features from the data with which to identify clear outliers (robust principle component analysis) \cite{RPCA}. This paper focuses on and compares against other machine-learning methods, including the expansion of the task from simply anomaly detection, to also applying semantic segmentation to the point cloud scene.

This task of open-set segmentation for point clouds has been explored previously \cite{real,doss}, using voxel-based architectures such as Cylinder3D \cite{cylinder3D}. One of the key challenges for this area is training a network when almost by definition, the training data should not have any examples of anomalous objects.  Cen et al \cite{real} approach this by using known instance labels in the training data to create synthetic anomalies during training, and demonstrate improved results on unknown (real) anomalies during testing.

As these existing methods however do not include anything like our reconstruction step, in our experiments we test our method against existing ones both with and without our reconstruction step added for a more thorough analysis. We also propose an alternative way to augment object instances which we demonstrate creates better synthetic anomalies for training. 

\subsection{Mamba for Point Cloud Scene Understanding}

\begin{figure}
    \centering
    \includegraphics[width=0.75\textwidth]{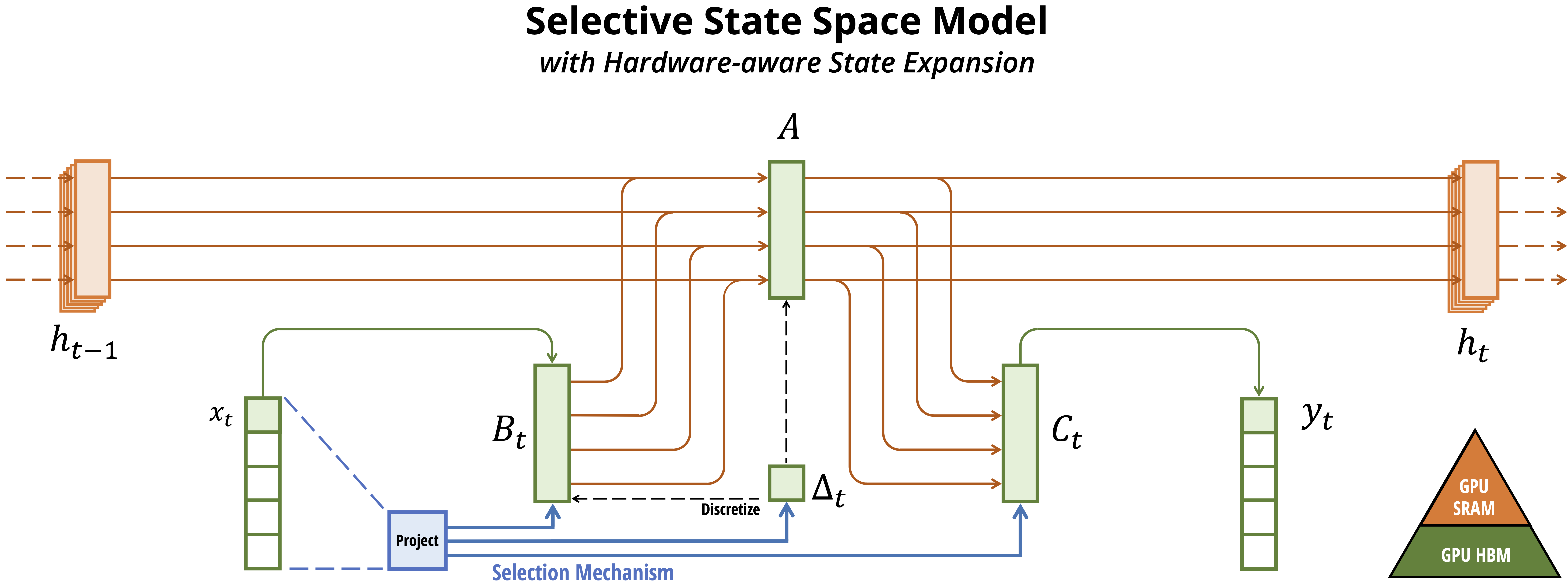}
    \caption[Mamba Block Diagram]{A visual diagram of the Selective State Space Model which forms the core of a Mamba block, from the Mamba paper \cite{mamba}.}
    \label{fig:mambaDiagram}
\end{figure}

Originating as a method for sequencing of dna, time, or large language models, Mamba \cite{mamba,mamba2} builds on the combined bases of Recurrent Neural Networks and State Space Models. It is able to store a ``hidden state'' of information (referred to as $h$ in \ref{fig:mambaDiagram}) while sequentially processing the data. The details of how much the hidden should both be updated, and be utilised, during the processing of each datapoint $x_t$, determined by applying learned weights to $x_t$, shown as blue lines in \ref{fig:mambaDiagram}. This process allows a network to selectively store only only useful information long-term while processing a sequence, while still being computationally efficient (with all computationally expensive steps capable of being performed in parallel on a GPU). Together this gives Mamba a powerful ability to model long-range dependencies, as well as efficient scalability to large datasets. For applications where data can be very large in scale such as outdoor-scene pointclouds, this makes it a very suitable architecture to use as a backbone.

Most computer-vision applications such as Vision Mamba \cite{VisionMamba} first flatten the data into a sequence, then apply each block in both directions (forwards and backwards) to allow the network to learn and apply long-range dependencies in both directions.

Applied to vision tasks ranging from those with causal data like video sequences \cite{VideoMamba,MambaMOS} to less causal ones like denoising or densification of static point clouds \cite{3DMambaIPF,SRMamba}, the architecture's effectiveness as a backbone, especially on large datasets such as LiDAR point clouds, has already been firmly established outside the context of anomaly detection.

Closed-set segmentation has also already shown success, either supplementing Mamba with ``K Nearest Neighbour''\cite{3D-UMamba,pointmambaKNNpaper} or Octree convolutions \cite{pointMamba} to extract close-range dependencies.

In the context of single-object, RGB image defect-detection, Mamba has been used previously \cite{MambaAD,ALMRR}. Two notable contrasts from this in our method are that we do not rely on pre-trained RGB networks such as ResNet (making our method more applicable to non-standard data types). We also investigate the challenging open-set, large-scale scene segmentation task instead of single-object defect detection.

\subsection{Reconstruction for Anomaly Detection}
Reconstruction as a preprocessing step is common for defect detection \cite{MambaAD} where a robust, pretrained, ResNet or similar can be used. Outside of this scope however, the use of reconstruction is rare. In examples where it has been used for open-set \cite{CoReSeg,GANReconstructFromSegmentation}, it uses predicted semantic labels as the reconstruction base. For example, Biase et al \cite{GANReconstructFromSegmentation} use a GAN to reconstruct a 2D image from their predicted semantic labels. Such a reconstruction will diverge from the input where there are inconsistencies in a given class (\eg cars being different colours), or errors in the semantic prediction being used as a base (\eg mislabeling one uncommon class for another). 

We argue it is more effective to train an autoencoder to encode and decode the original input, removing any reliance on predicted semantic labels. This allows the autoencoder to learn to reconstruct a scene's ``default context'', directly assisting with outlier detection. This is inspired by reconstruction steps in single-object defect-detection which will fail to recreate scratches, blemishes, or other defects, aiding in their identification.

\section{Method}

\begin{figure}[htbp]
  \centering
  \includegraphics[width=0.8\textwidth]{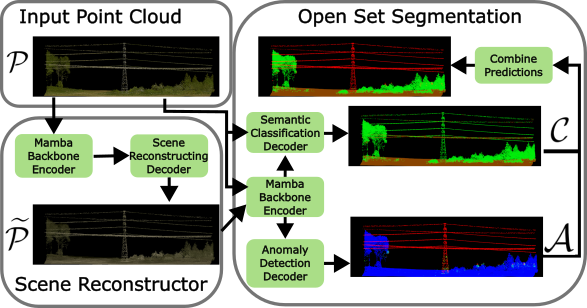}
  \caption{High level diagram of our overall method for inference. Our scene reconstructor can be separated out to assist any existing anomaly detection architecture, not just our MambaAD. Red points are anomalous, brown are groud, green is vegetation. For anomaly vs in-set, blue points are in-set.}
  \label{fig:HighLevelDiagram}
\end{figure}

Our two separate architectures are shown in figure \ref{fig:HighLevelDiagram}. We define each point cloud $\cP$, as having an (inconsistent) number of points $N$, such that $\cP=\{ \bp_i \}^{N}_{i=1}$. First, a learned reconstruction of the scene $\widetilde{\cP}=\{ \widetilde{\bp_i} \}^{N}_{i=1}$ is generated. We find relying on the inconsistencies between this reconstruction and the original values (RGB, Intensity, XYZ), improves performance, especially for detection of out-of-set anomalies. 

The final result for the task of open-set segmentation is defined as two sets of predicted labels, the semantic class prediction $\cC$ where $\cC=\{ \bc_i \}^{N}_{i=1}$, and a separate anomaly score $\cA$, where $\cA=\{ \ba_i \}^{N}_{i=1}$. The class prediction is an integer from 0 to $k$, where $k$ is the number of classes in the training data, excluding any anomaly classes (real or synthetic) used to train for anomaly detection. The anomaly score is a single value from 0 to 1. We provide a Mamba-based anomaly detection architecture, in contrast to convolutional backbones such as the  Cylinder3D\cite{cylinder3D}, used by REAL\cite{real} and DOSS\cite{doss}. Our architecture generates for a point cloud $\cP$ both semantic class predictions $\cC$ , as well as an anomaly score $\cA$.

The groundtruth class labels for each point are $\cG$ where $\cG=\{ \bg_i \}^{N}_{i=1}$. In training and testing these labels are mapped and masked to either $\cG_A$ or $\cG_c$ to be used as the groundtruth for anomaly detection or semantic class prediction respectively.

To ensure both robust performance as well as scientifically rigorous experiments, the training data undergoes masking of three different types. Unknown objects to be used during testing must be completely hidden during training, the removal of such points is defined as $k()$ such that $\cP_{known} = k(\cP)$. For training the semantic segmentation, the ``known'' anomalous points (either synthetic anomalies for KITTI or a defined subset of anomalies in ECLAIR) are also removed to create an anomaly-free scene we refer to as the ``default context''. This process defined as $d()$, $\cP_{default} = d(\cP)$.

\subsubsection{Mamba Encoder Backbone}

\begin{figure}[htbp]
  \centering
  \includegraphics[width=0.99\textwidth]{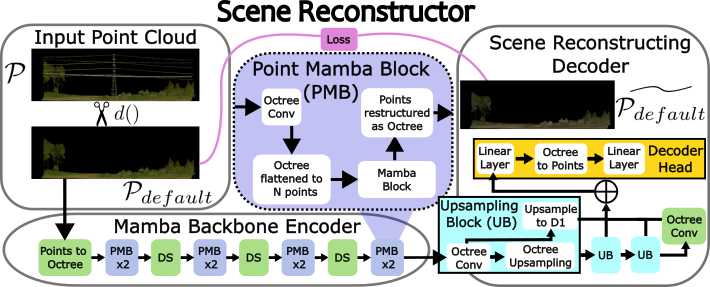}
  \caption{Scene Reconstructor architecture for training. In this example the transmission tower and it's power lines are considered anomalous objects}
  \label{fig:SceneReconstructor}
\end{figure}

For the backbone, we use a modified version of "Point Mamba" created by Liu et al\cite{pointMamba}. As shown in the bottom left of \ref{fig:SceneReconstructor}, it uses multiple Mamba blocks between each downsampling to effectively capture global scene information, while \textit{also} using octree convolutions to utilise the more local information around each point. 

\begin{figure}[htbp]
  \centering
  \includegraphics[width=0.75\textwidth]{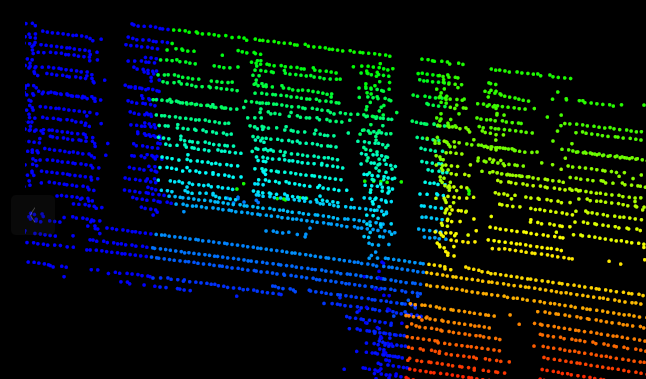}
  \caption{A wall with points coloured according to position in flattened sequence. One XY block ascends from blue to green, and the next descends, green to red.}
  \label{fig:PointFlattening}
\end{figure}

Details for the backbone are available in the original paper \cite{pointMamba}, we make two modifications, the first is updating the Mamba blocks with the more recent Mamba-2 \cite{mamba2}. The second is changing the flattening process used to turn 3D data into the one directional sequence a Mamba block can be applied to. Instead of a simple three-dimensional z-order curve as used in prior 3D Mamba architectures, we used a multi step process to improve the grouping of points which belong to the same object instance. 

First we sort points into 2 metre square blocks along the XY place (ground plane). Blocks are sorted using a Z-order curve as usual, and within each block points are sorted by their Z-axis value. The direction of the Z-axis sorting alternates between each adjacent block in an effort to keep points belonging to the same object instance, together in the final sequence. In figure \ref{fig:PointFlattening} there a wall which is spread across two XY blocks, has a continuous sequence of points.

\subsection{Learned Scene Reconstruction}

Following from successes in single-object anomaly detection, as well as a small number of scene-based open-set segmentation, we encode and decode point cloud scenes to better identify out-of-distribution anomalies. The Mamba backbone is used to encode an information-dense version of the scene $\{ \be_i \}^{N_1}_{i=1}$. We then train a decoder to reconstruct the scene from this. The decoder uses octree convolutions between each upsampling, followed by a series of linear layers. Together, this creates the Mamba-based autoencoder shown in figure \ref{fig:SceneReconstructor}, which reconstructs a given scene $\cP$, as $\widetilde{\cP}$. The training loss for this reconstructor $\L_R$ is simply $L_R = CrossEntropy(\cP,\widetilde{\cP})$, with both point clouds masked to exclude anomalous objects.

\subsubsection{A Scene's Default Context}
We identify anomalies using both the absolute and signed difference between the original and reconstructed point clouds: $\Delta\cP = \{\cP - \widetilde{\cP},\| \cP - \widetilde{\cP} \|\}$. With this input, both our anomaly detection architecture and others demonstrate improved performance.

Our logic is that by learning the features of the known objects in a scene, referred to as the "default context" in the training data, the reconstructor will fail more, and by larger margins at reconstructing outlier objects (anomalies), aiding in their identification. The second half of our architecture then uses this $\Delta\cP$ both to effectively segment the anomalous objects, identifying the needles in the haystack so to speak.

Another notable benefit from relying on the reconstructed feature difference $\Delta\cP$ is that it abstracts the problem of anomaly detection away from the raw data values, leading to less overfitting on the known training data. For this task, the training data is often either a small selection of ``known'' anomalies, or synthetic anomalies, both of which can easily lead to overfitting issues when new anomalies in testing or application are completely differently in appearance.

For a problem task which revolves around robust and reliable networks which work on completely unknown data, this is a significant advantage.

\subsubsection{Reconstructed Features}
In addition to any available raw data features (RGB colour information, intensity/reflectance), we reconstruct the geometric information as well. As the point clouds can be as large in the XY plane as a 100x100 metre square, simply reconstructing the XYZ information directly is very challenging to do with accuracy. If normalised to a value from -1 to 1 for example, a slight deviation in decimals can lead to huge changes in the 100m scene, making errors too large to be of use.

We instead reconstruct the XYZ of each point relative to it's Octree leaf node. Each point $\bp$ will sit inside a given leaf node of the octree $n$, where the middle of that octree node is the co-ordinate $x_n,y_n,z_n$. The relative $x$ co-ordinate $x_r$ is a simple calculation of $x_r = x_p - x_n$ for the x-axis, and the same for $y_r$ and $z_r$. By reconstructing this instead of the global position $x_p$, objects are still in the correct location, and errors instead present as malformed surfaces and exteriors, \eg a flat surface no longer being flat, or a more three-dimensional object such as a bush becoming flatter, aiding in the identification of anomalies.

\subsection{Mamba-Based Anomaly Detection}

While Mamba has been used previously for the task of single-object anomaly detection (identifying defects)\cite{MambaAD}, we are the first to apply a Mamba backbone to the task of anomaly detection in scenes (open-set segmentation).

Using this backbone, we then have two separate decoder heads, one for the anomaly detection score $\cA$, and one for semantic segmentation of known, in-set objects $\cC$.

\begin{figure}[htbp]
  \centering
  \includegraphics[width=0.99\textwidth]{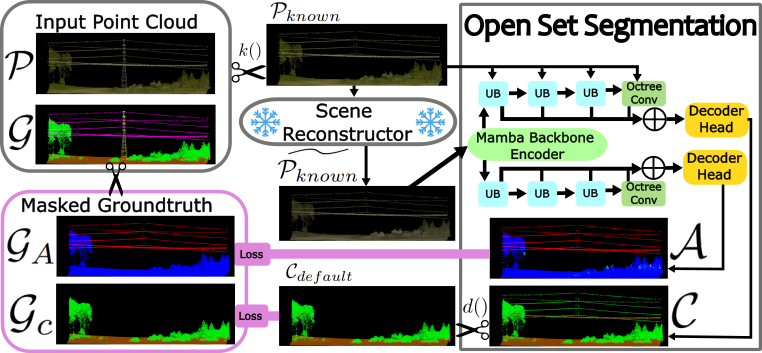}
  \caption{Our Open Set Segmentation architecture during training.}
  \label{fig:OpenSetSegmentation}
\end{figure}

Our loss is calculated using both of these predictions and their respective groundtruth values $\cG_C$ and $\cG_A$, as visualised in figure \ref{fig:OpenSetSegmentation}. The final loss is $L = L_A + L_C$ where $L_A = BinaryCrossEntropy(\cA,\cG_A)$ and $L_C = Cross Entropy(\cC,\cG_C)$.

Anomalies ``seen'' during training (either synthetic, or a subset of the anomalous classes) often vary greatly from anomalies encountered in practice, making overfitting a significant issue. Known objects however, assuming reasonable training sample size, are more likely to match the training data. We therefore include skip connections in the semantic decoder head, but not the anomaly detection head where they can make the network more likely to overfit to the "training data anomalies" only.

When incorporating the reconstruction difference $\Delta\cP$, the original point data is still extremely important for the task of open-set segmentation. While errors in the reconstruction are well-suited to anomaly detection, they provide little useful information for semantic segmentation of known classes. To this end, we add the raw data back in at each layer of the semantic segmentation decoder, as shown in figure \ref{fig:OpenSetSegmentation} where $P_known$ is repeatedly passed in to the open set segmentation model's upsampling blocks (UB) for the semantic prediction but not the anomaly detection one.

\subsection{Creating Synthetic Anomalies}
As this is the task of identifying objects for which we lack training data, previous works have already explored creating synthetic data instead. We follow from Cen et al's REAL \cite{real} in assuming the training data has not just semantic labels but also object instance labels for each point, to then randomly augment known objects, creating synthetic ones.

They use a scaling based method, changing the size of the object, however we are concerned that this simply leads to anomaly detection networks learning to either identify point density disparity (objects made larger are sparser than their surroundings, objects made smaller are more point-dense), or floating objects / objects which exist below the ground plane, as the object's base becomes lower or higher in the Z axis due to the overall size change.

As an alternative, we instead use an augmentation inspired by the ``Rubik's Cube'' toy. Each object instance is split down the middle (X,Y,Z) into eighths. Different halves of the object are then randomly rotated along their axis to create a synthetic anomalous object, which does not break the overall continuity of the scene (whether point density or the ground plane). 

By only rotating each side like a Rubik's cube, as opposed to randomly flipping the individual eigths, we ensure that the corner of each eighth originally touching the object's centrepoint, continues to do so. This is to minimise the chance of \eg accidentally creating multiple separate objects, with the assumption that few object instances are empty in their centre.

\section{Experiments}
We ran experiments on both the Semantic KITTI \cite{semanticKitti} and ECLAIR \cite{eclair2024} datasets. We thoroughly test our reconstruction and anomaly detection methods both together and separately, against existing methods \cite{real,doss}. These methods were run using their preset settings, while the hyperparemeter specifics of our method are available in our code.

We will note here decisions made for the experimental setup that differ to previous works in this area: 

As anomalies used for testing should not benefit the training whatsoever, we do not simply set their weighting to 0 when calculating the loss. We remove anomalous objects completely before training, referred to in our method as $k()$. This is to ensure the network cannot benefit even indirectly, such as from learning to identify the borders of objects adjacent to an anomaly, for which the loss is not ignored.

What to use as anomalous objects in training to calculate loss, is also a difficult task. One solution has been to simply group all anomalous objects as one ``unknown class'' and use it to train a separate loss for anomaly detection \cite{doss}, however we feel this leads to difficulty in knowing whether results would be replicated on anomalous objects which are truly unseen during training. In the context of semantic kitti for example, there is only one anomalous object, ``other-vehicle'', available for both training and testing. A network trained using ``other-vehicle'' would only need to learn to identify this specific class to perform well in testing, not to identify anomalies in general.

If object instance labels are available, we agree with Cen et al \cite{real} that synthetic anomalies are a promising solution. In KITTI where this is available, we test both their scaling-based synthesis, and our novel rubiks-cube inspired approach.

In ECLAIR where instance labels are unavailable, we opt to use half the anomalous classes for training (referred to as ``known'' anomalies) and keep half exclusively for testing. This enables the performance on truly unknown anomalies to still be inferred from the experiments.

\subsection{KITTI Dataset}

\begin{table}
    \begin{tabular}{ |p{5cm}|p{2.2cm}||p{1.3cm}|p{1.3cm}||p{1cm}|}
     \hline
     \multicolumn{5}{|c|}{Semantic KITTI Dataset} \\
     \hline
     Architecture & Augmentation & AUROC & AUPR & MIoU\\
     \hline
     REAL (quoted from paper) & Scale & 84.9 & 20.8 & \textbf{57.8} \\
     REAL  & Rubiks & 84.6 & 6.3 & 46.8 \\
     Mamba AD  & Scale & 83.74 & 2.7 & 30.7 \\
     Mamba AD  & Rubiks & 93.3 & 12.2 & 24.8 \\
     \hline
     \multicolumn{5}{|c|}{ \textsuperscript{\dag} Using our reconstruction step trained on default scenes} \\
     \hline
     REAL \textsuperscript{\dag} & Rubiks & 83.6 & 2.0 & 24.0 \\ 
     Mamba AD \textsuperscript{\dag} & Scale & 82.3 & 3.0 & 22.0 \\
     Mamba AD \textsuperscript{\dag} & Rubiks & 95.6 & 25.4 & 24.4 \\
     \hline
     \multicolumn{5}{|c|}{ \textsuperscript{\ddag} Using our reconstruction step trained on objects only} \\
     \hline
     REAL\textsuperscript{\ddag} & Rubiks & 78.0 & 1.4 & 23.5 \\ 
     Mamba AD \textsuperscript{\ddag} & Scale & 85.5  & 3.15 & 23.4 \\
     Mamba AD \textsuperscript{\ddag} & Rubiks & \textbf{96.3} & \textbf{37.5}  & 25.8 \\
     \hline
    \end{tabular}
    \caption{Semantic KITTI Results}
    \label{tab:SemKitti}
\end{table}

\begin{figure}
    \centering
    \subfloat[a][Original Scan]
    {
        \includegraphics[width=0.33\textwidth]{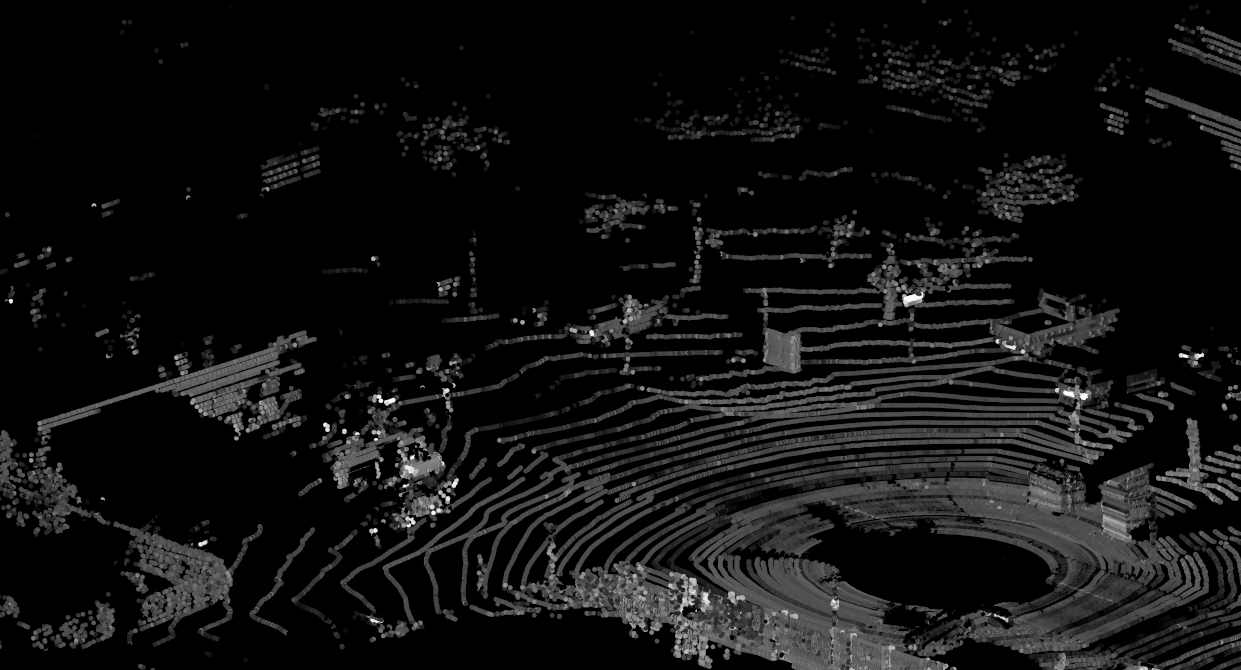}
    }
    \subfloat[d][Mamba AD]
    {
        \includegraphics[width=0.33\textwidth]{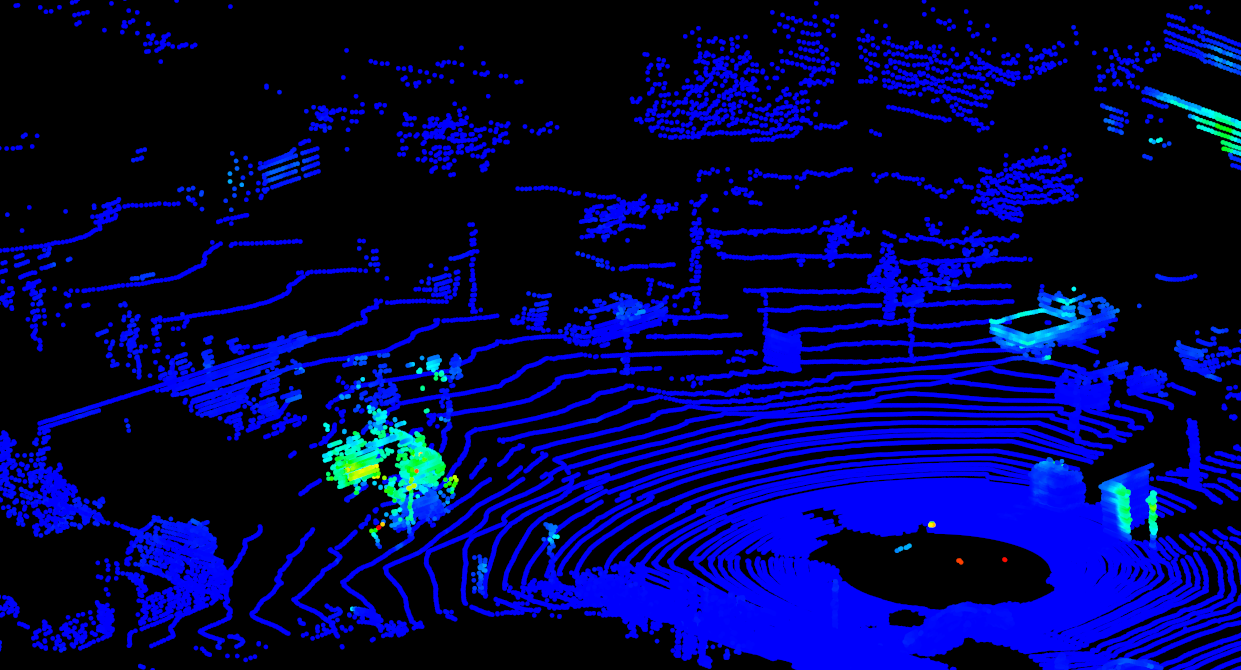}
    }
    \subfloat[c][REAL]
    {
        \includegraphics[width=0.33\textwidth]{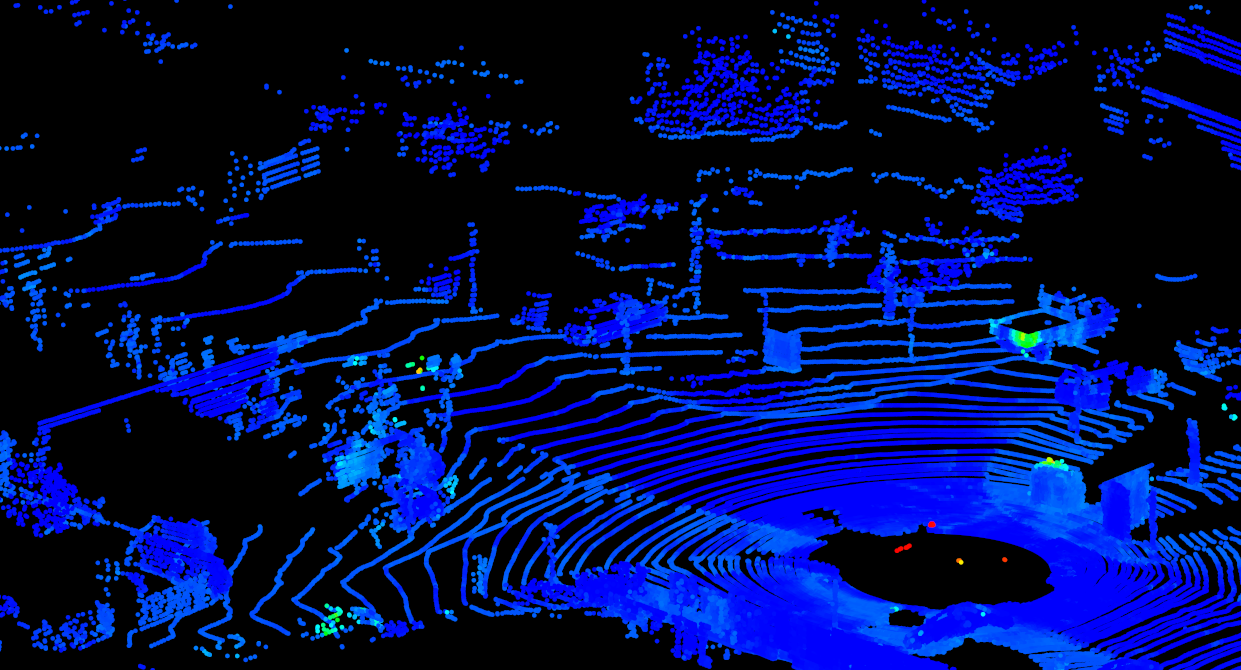}
    }\\
    \subfloat[d][Groundtruth]
    {
        \includegraphics[width=0.33\textwidth]{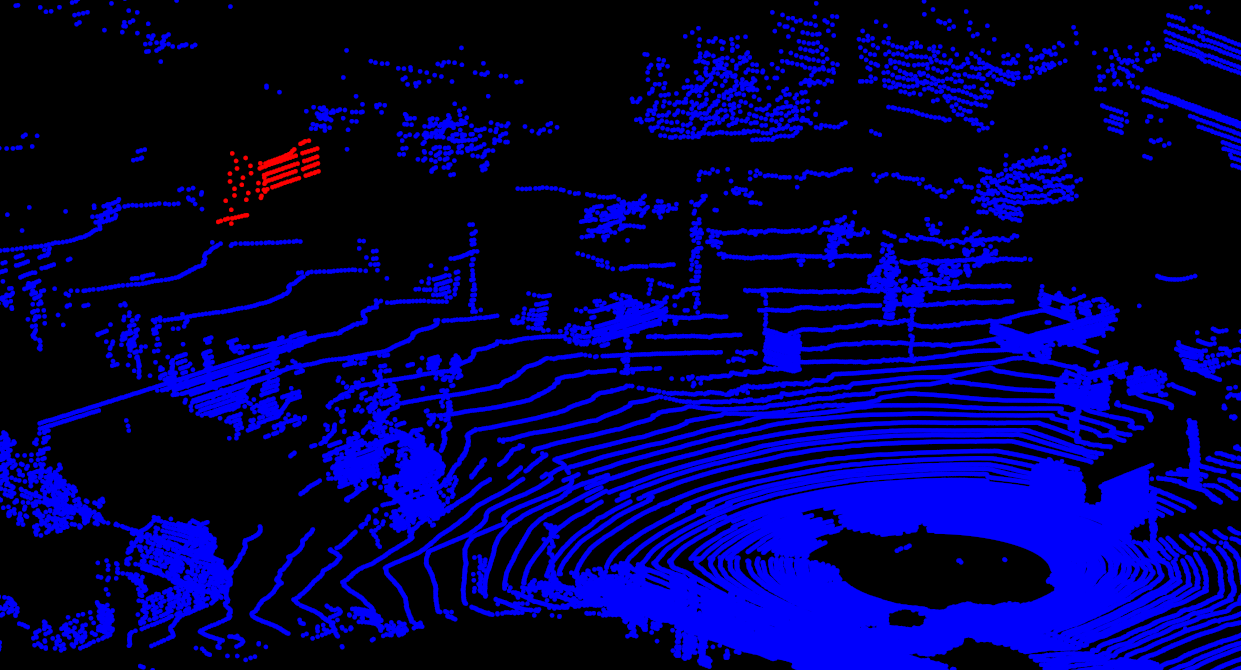}
    }
    \subfloat[e][Mamba AD\textsuperscript{\dag}]
    {
        \includegraphics[width=0.33\textwidth]{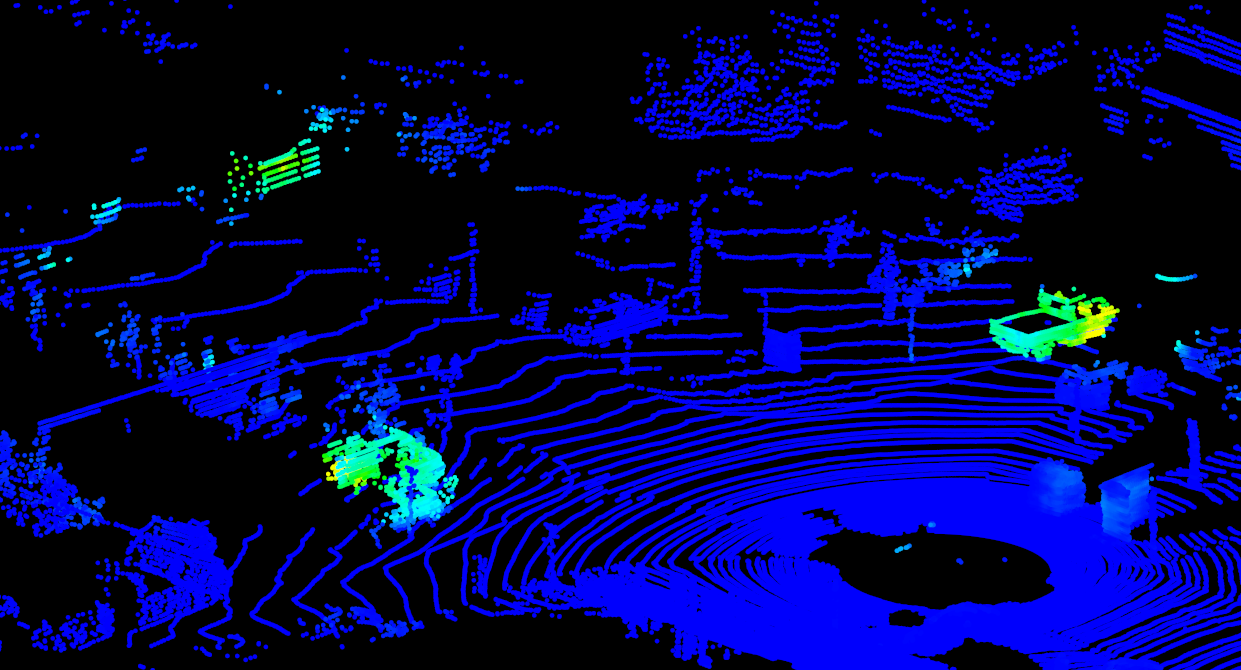}
    }
    \subfloat[f][Mamba AD\textsuperscript{\ddag}]
    {
        \includegraphics[width=0.33\textwidth]{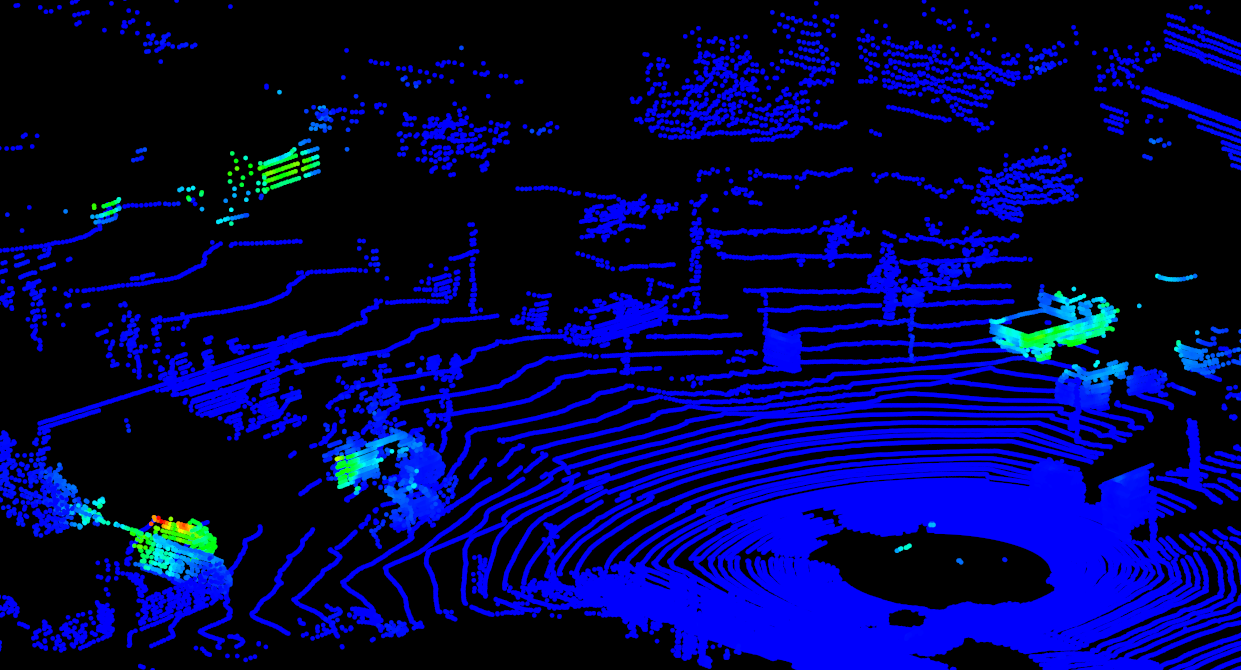}
    }
    \caption{An anomaly detection example in KITTI. The single anomalous vehicle is extremely difficult wto identify from only a single static scan.
    \newline\textsuperscript{\dag} is using our reconstruction step trained on all known objects. \newline{}\textsuperscript{\ddag} is using our reconstruction step trained only on object instances.}
    \label{fig:KITTIQualitative}    
\end{figure}
Following with previous works \cite{doss,real}, we use Semantic Kitti to test anomaly detection and open-set segmentation, setting ``other-vehicle'' as the anomalous class, with all other classes treated as ``known objects''. For REAL's original unaltered method (no reconstruction step, scaling based augmentation), we quote their results directly from their paper. As DOSS uses the same anomaly used for testing (``other-vehicle'') to calculate it's Objectosphere Loss during training, we do not include it in these experiments.

As shown in table \ref{tab:SemKitti} our Mamba AD method achieves superior performance in anomaly detection, for both the metrics of Area under the Receiver Operating Characteristic Curve (AUROC) and Area under the Precision Recall Curve (AUPR). AUPR is much more volatile and unreliable on datasets with significant imbalances between positive and negative, but we include it as well for good measure. Mean Intersection over Union (MIoU) is used to measure the semantic segmentation performance for known classes. The improved performance with our Rubiks cube based augmentation compared to Scaling, suggests that the scaling based approach does lead to problems such as the network learning to focus on point density or relation to ground plane. This perhaps enables the open-set segmentation network, having learnt a "shortcut trick", to then devote more weights to the problem of segmentation, which would explain the drop in MIoU on known objects from introducing Rubiks-cube anomalies. Finally, training the reconstruction step on objects only, more akin to the approach of single-object defect detection, shows improvement for our method, but not for REAL. 

We include an example of what this challenge looks like with a static point cloud in figure \ref{fig:KITTIQualitative}, where there is one ``other-vehicle'', displayed as red in the groundtruth, with the anomaly prediction score of each method shown as a heatmap.

\subsection{ECLAIR Dataset}

ECLAIR is a large-scale outdoor scene aerial LiDAR dataset, with each scan being a hundred square metres. We chose to test both our method, and two comparable ones (REAL, DOSS) on ECLAIR. We chose ECLAIR and not, \eg Nuscenes as it more different from KITTI, being rural scenes instead of urban, and more importantly, large in both number of points and physical scene size, allowing the scalability of methods (\eg Mamba vs Cylindrical Voxel Backbone Architectures) to be properly tested. We use the reccomended train, validation, and test splits provided by the dataset. For REAL we also included testing using it's ``Predictive Distribution Calibration'' approach which does not require any anomalies during training. 

As both REAL and DOSS use Cylinder3D as a backbone, we had to modify the cylinder feature generator from a series of linear layers increasing the feature size of each point, followed by a max pooling of points, to do the max pooling when the feature size is smaller, with linear layers afterwards to reach the desired feature size. Without this modification the architectures simply could not handle large numbers of points such as is common in ECLAIR. They were otherwise kept the same. Our Mamba backbone required no changes compared to the KITTI experiments.

The ``known anomalies'' used for training on ECLAIR are ``Trans Wires'', ``Dist Wires'', ``Dist Poles'', and ``Vehicles''. The ``unknown anomalies'' for testing are ``Unassigned'', ``Buildings'', ``Noise'', ``Trans Towers'', and ``Fence''. The groupings were chosen based on balancing the proportion of points in the dataset as close to 50/50 as possible, and ensuring similar objects were grouped together (Trans Wires and Dist Wires). The ``default scene'' used to train the reconstruction step includes neither the known nor unknown anomalies.

\begin{table}
    \begin{tabular}{ |p{2.8cm}|p{1.5cm}||p{1.25cm}|p{1.25cm}||p{1.6cm}|p{1.6cm}||p{0.8cm}|  }
     \hline
     \multicolumn{7}{|c|}{ECLAIR Dataset} \\
     \hline
     Architecture & Anomalies in \newline training &AUROC \newline (Known)& AUPR \newline (Known) & AUROC \newline (Unknown) & AUPR \newline (Unknown) & MIoU\\
     \hline
     REAL   & \xmark & 77.5 & 1.10 & 76.0 & 0.5 &  69.0\\
     REAL   & \cmark  & 97.6 & 1.19 & 90.5 & 1.4 & 90.6 \\
     DOSS   & \cmark  & 74.4  & 3.5 & 65.7 & 0.3 & 86.5 \\
     Mamba AD   & \cmark& 99.96 & 99.2 & 87.7 & 18.1 & \textbf{96.9}\\
     \hline 
    \multicolumn{7}{|c|}{\textsuperscript{\dag} Methods using our reconstruction step as preprocessing} \\
     \hline
     REAL\textsuperscript{\dag}   & \cmark & 99.1 & 39.7 & \textbf{91.1} & 2.3 & 92.9 \\
     DOSS\textsuperscript{\dag}   & \cmark & 82.0 & 10.7 & 64.7 & 0.4 & 89.2\\
     Mamba AD\textsuperscript{\dag}  & \cmark & \textbf{99.97} & \textbf{99.5} & 88.8
     & \textbf{34.6} & 96.6\\
     \hline
    \end{tabular}
    \caption{Eclair Results}
    \label{tab:Eclair}
\end{table}

In table \ref{tab:Eclair} we can see the benefits of the Mamba backbones in full. Both the utilisation of long-range dependencies and scalability to large data. Our Mamba AD method outperforms on all metrics except AUROC of unknown classes, for which our reconstruction step still shows improvement for the existing method REAL. On the ``Known'' anomalies we achieve near perfect performance, however this near-perfection does not carry over to the unknown anomalies. We consider this an example of why it's important that anomaly detection is tested on different \textit{object classes} than used in training, not just different LiDAR scans. The low performance by REAL without any anomalies during training shows the importance of having some form of anomaly to calculate loss from, whether synthetic or otherwise.

\begin{figure}[h!]
    \centering
    \subfloat[a][Groundtruth]
    {
        \includegraphics[width=0.25\textwidth]{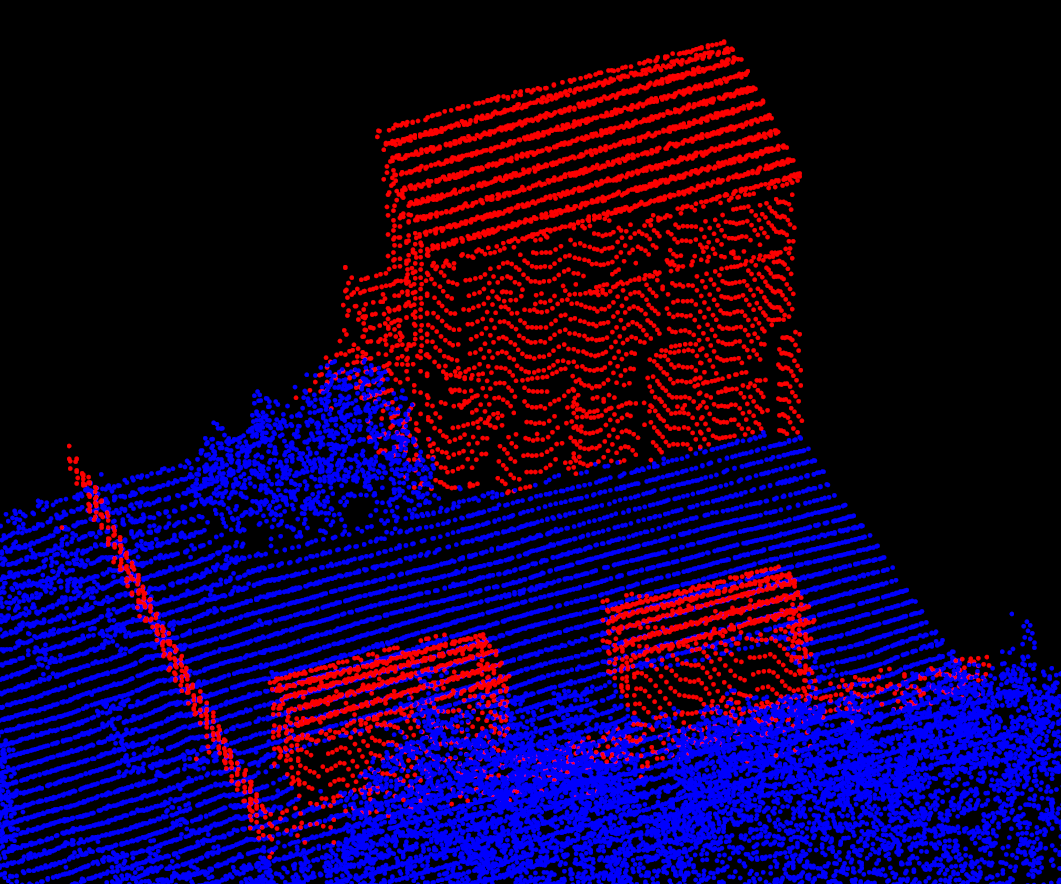}
    }
    \subfloat[b][Mamba AD\textsuperscript{\dag}]
    {
        \includegraphics[width=0.25\textwidth]{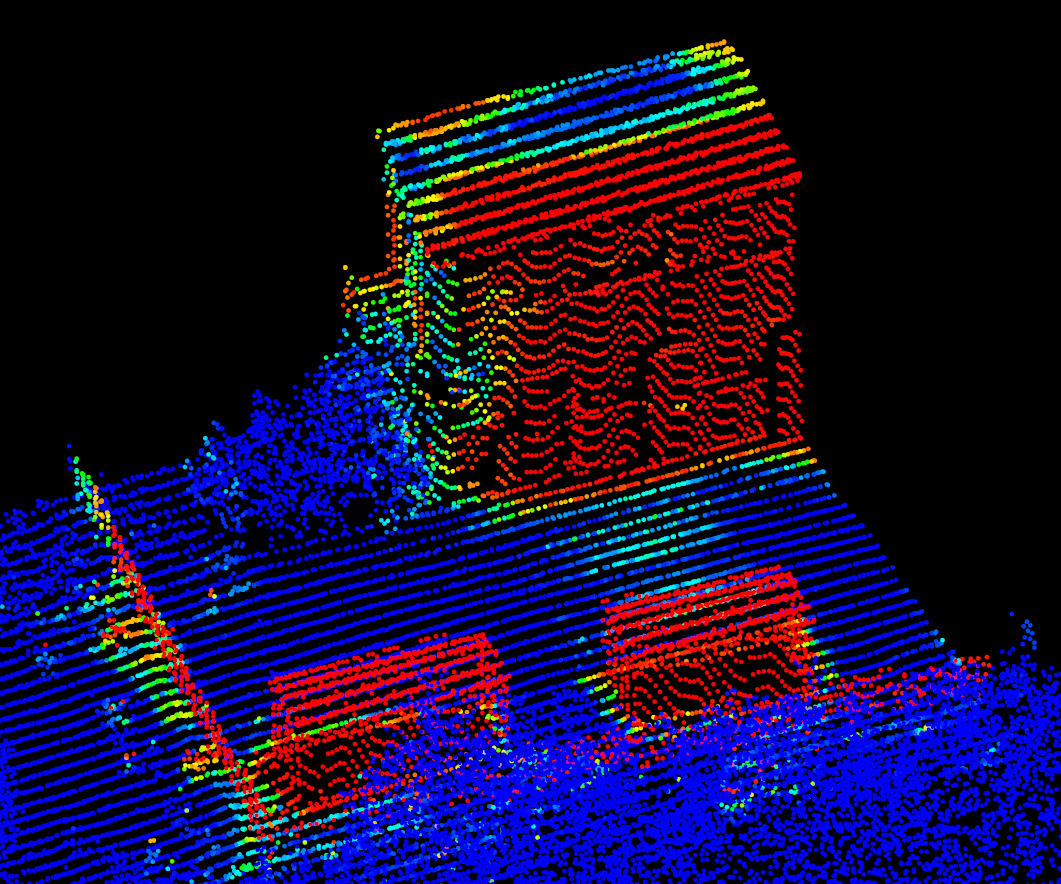}
     }
    \subfloat[c][REAL\textsuperscript{\dag}]
    {
        \includegraphics[width=0.25\textwidth]{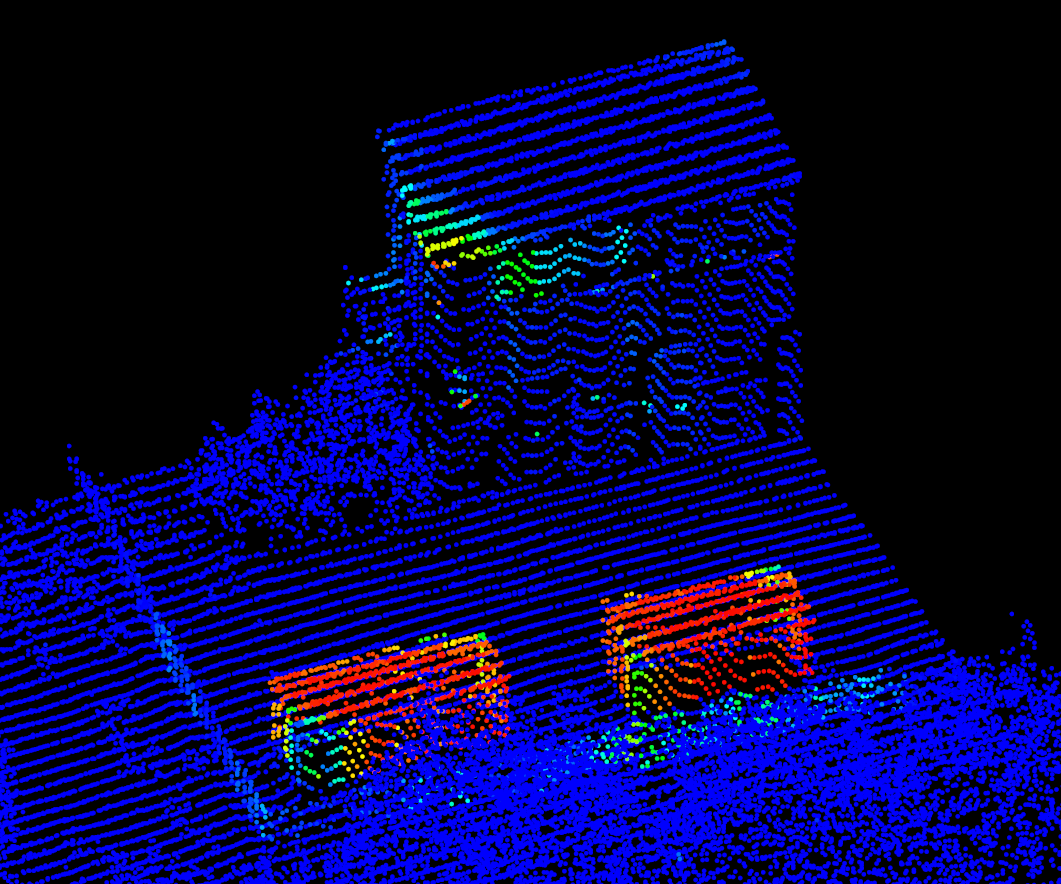}
    }\\
    \subfloat[d][Groundtruth]
    {
        \includegraphics[width=0.25\textwidth]{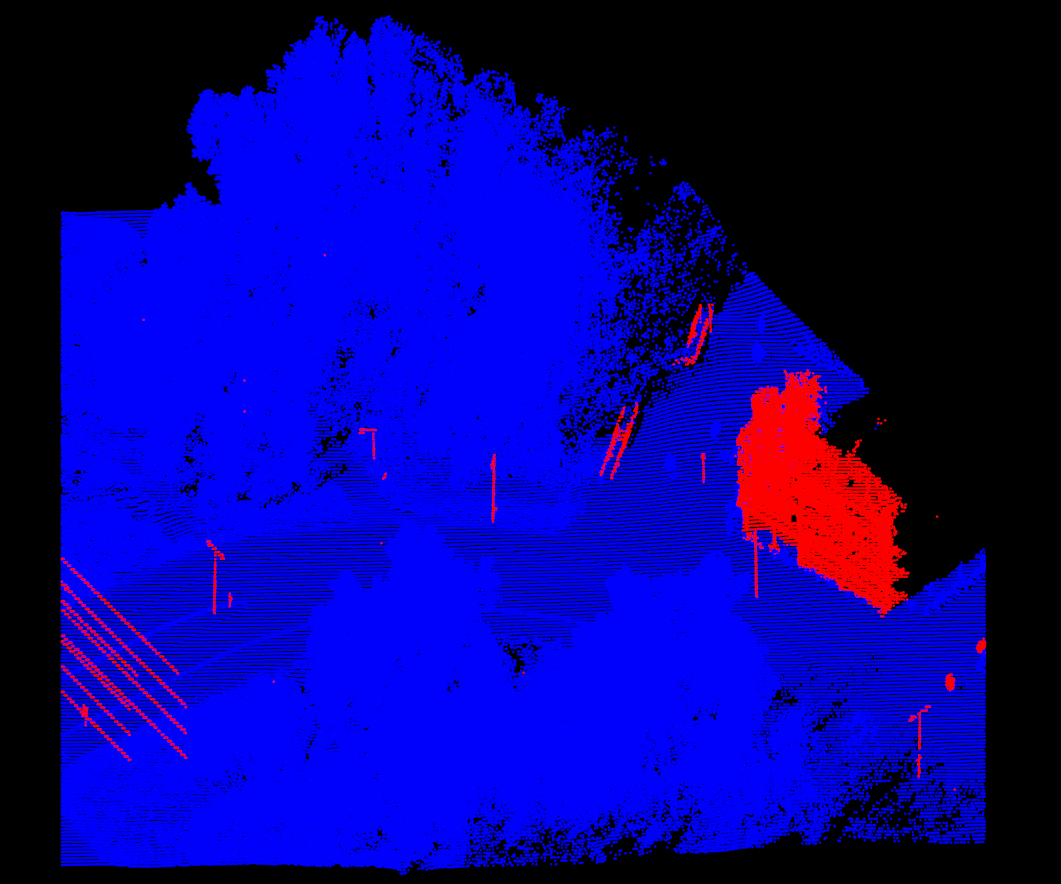}
    }
    \subfloat[e][Mamba AD\textsuperscript{\dag}]
    {
        \includegraphics[width=0.25\textwidth]{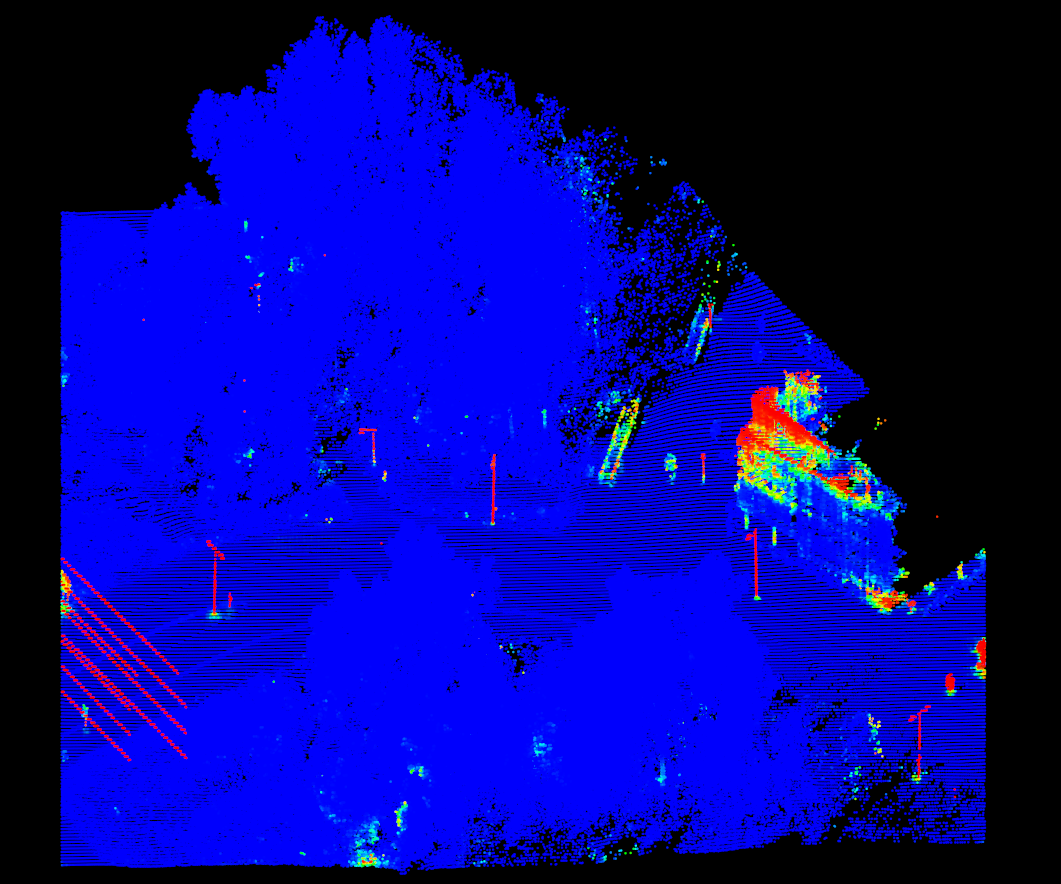}
    }
    \subfloat[f][REAL\textsuperscript{\dag}]
    {
        \includegraphics[width=0.25\textwidth]{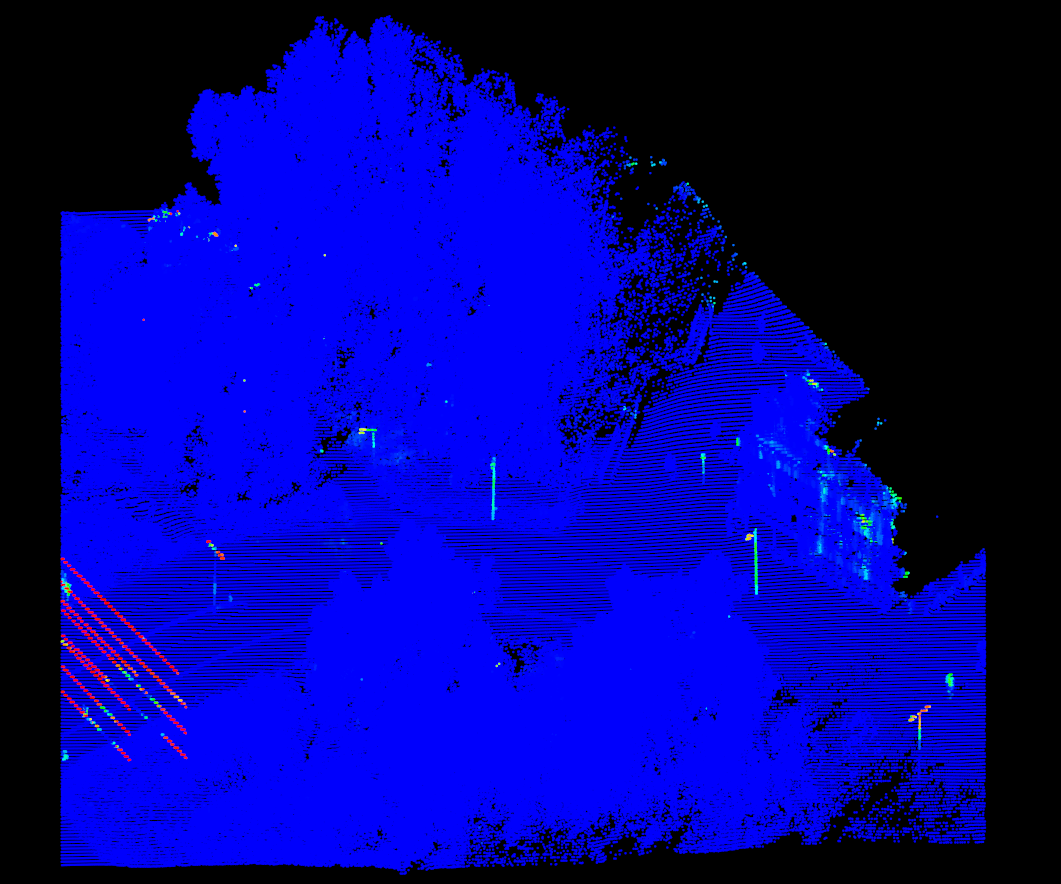}
    }
    \caption{Challenging anomaly detection examples. \textsuperscript{\dag} Using our reconstruction step.}
    \label{fig:EclairADQualitatitve}    
\end{figure}

\begin{figure}[h]
    \centering
    \subfloat[a][Original Scan]
    {
        \includegraphics[width=0.33\textwidth]{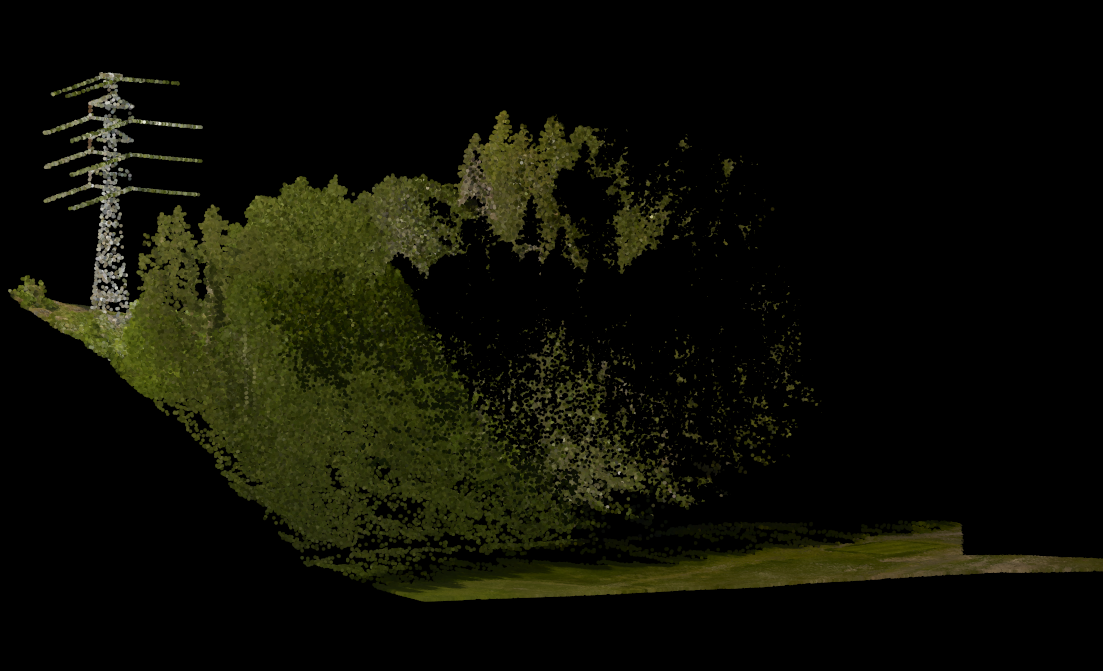}
        
    }
    \subfloat[b][Groundtruth Labels]
    {
        \includegraphics[width=0.33\textwidth]{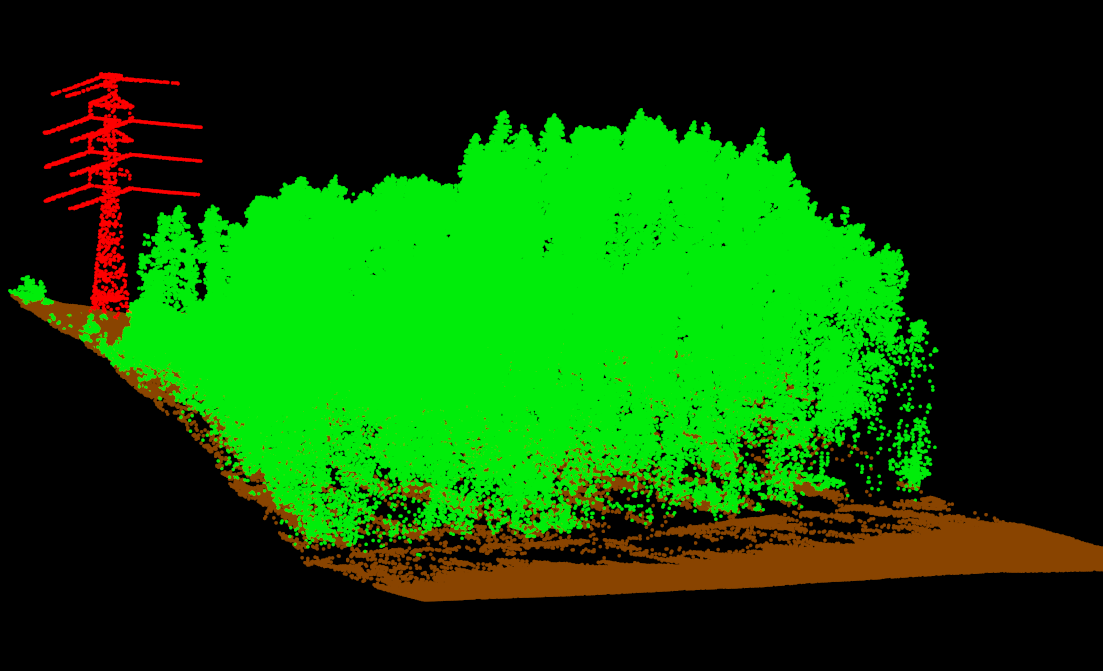}
    }
    \subfloat[c][REAL (limited training)]
    {
        \includegraphics[width=0.33\textwidth]{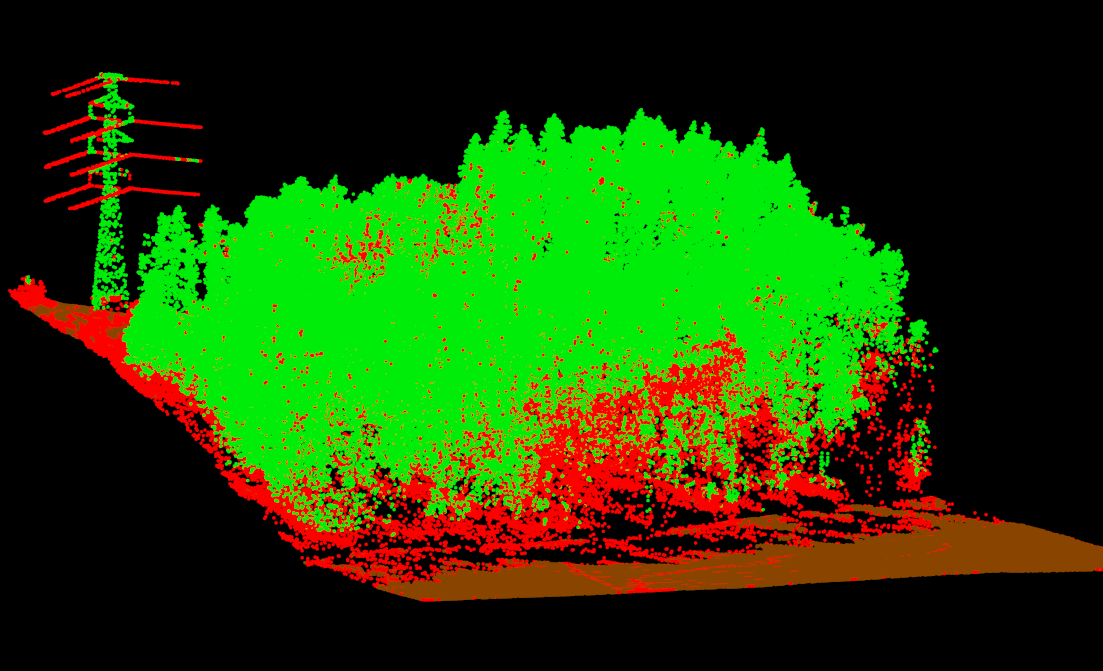}
    }\\
    \subfloat[d][Mamba AD]
    {
        \includegraphics[width=0.33\textwidth]{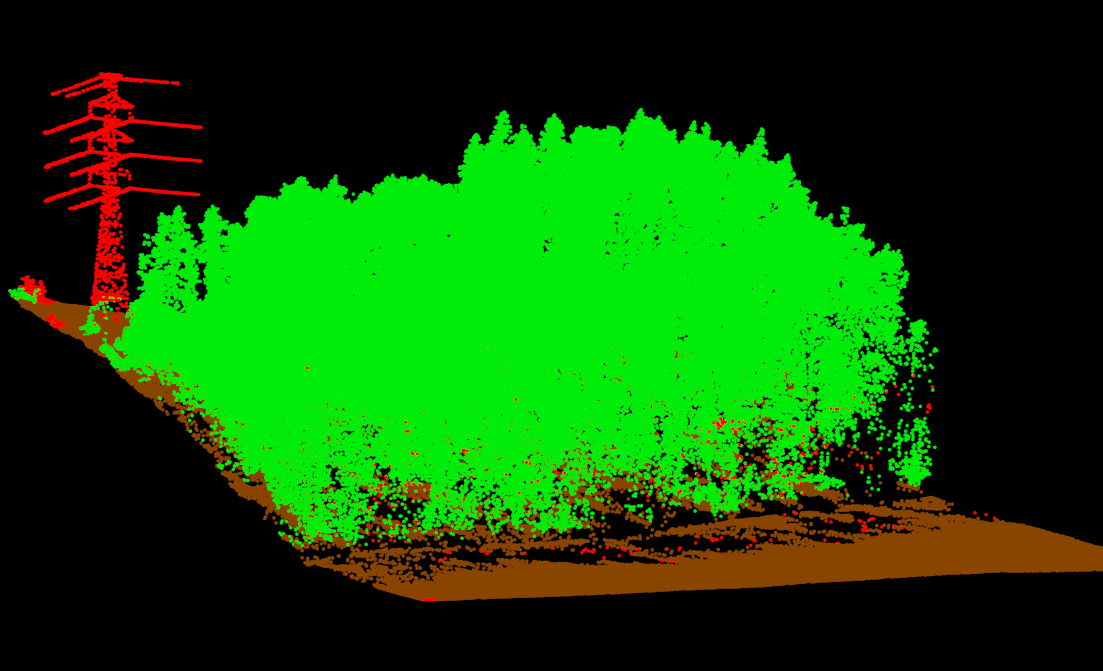}
    }
    \subfloat[e][DOSS]
    {
        \includegraphics[width=0.33\textwidth]{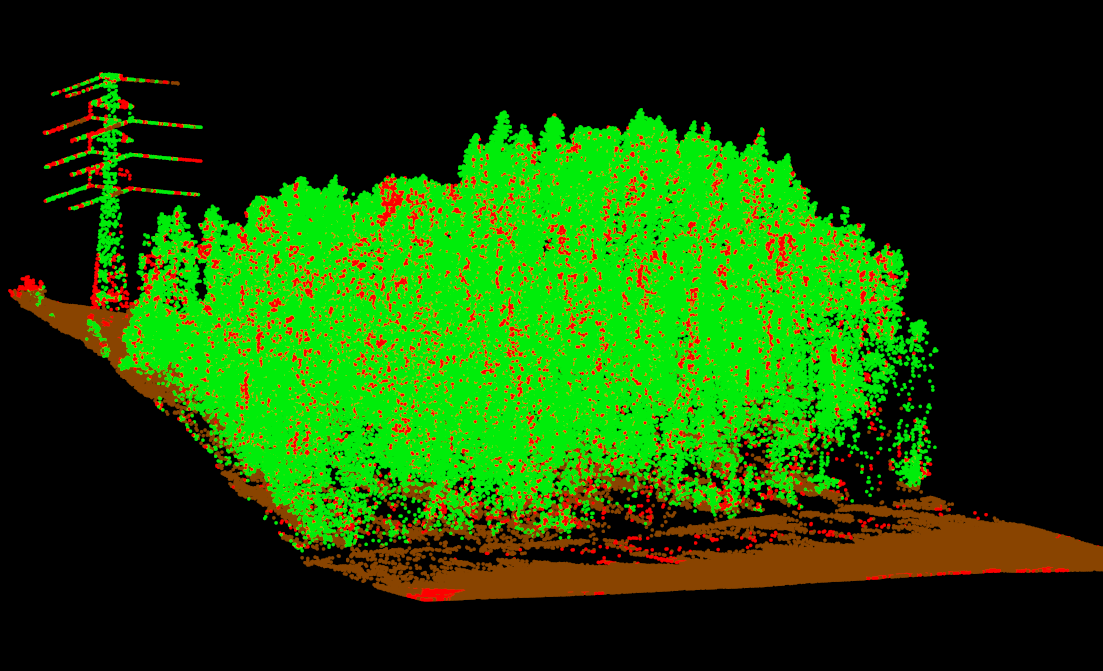}
    }
    \subfloat[f][REAL]
    {
        \includegraphics[width=0.33\textwidth]{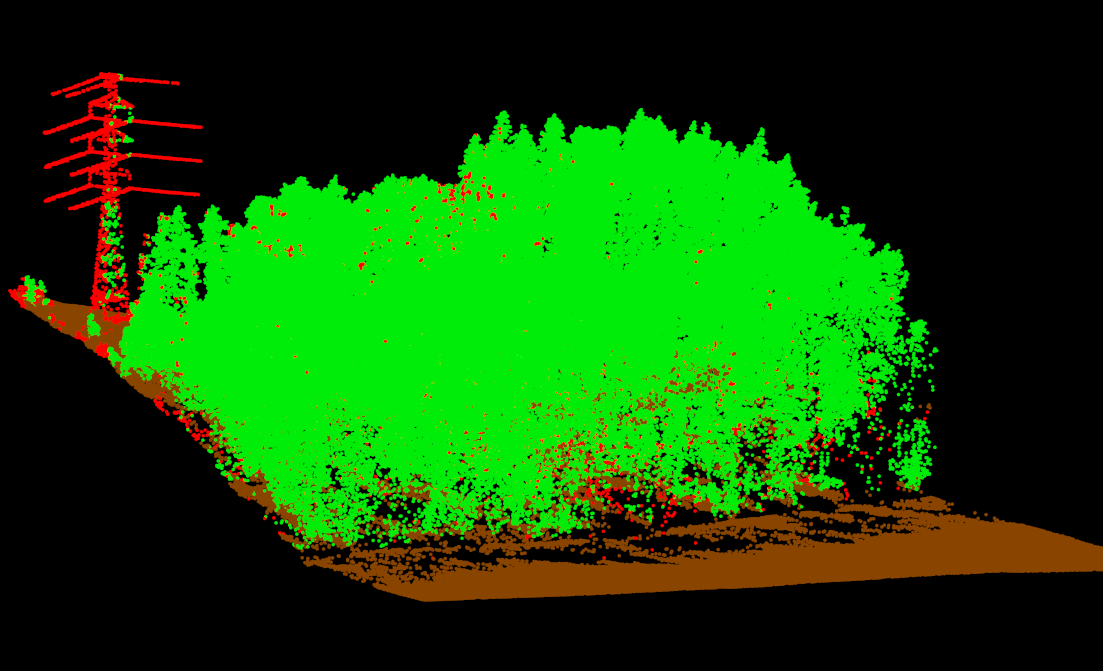}
    }\\
    \subfloat[g][MAMBA AD\textsuperscript{\dag}]
    {
        \includegraphics[width=0.33\textwidth]{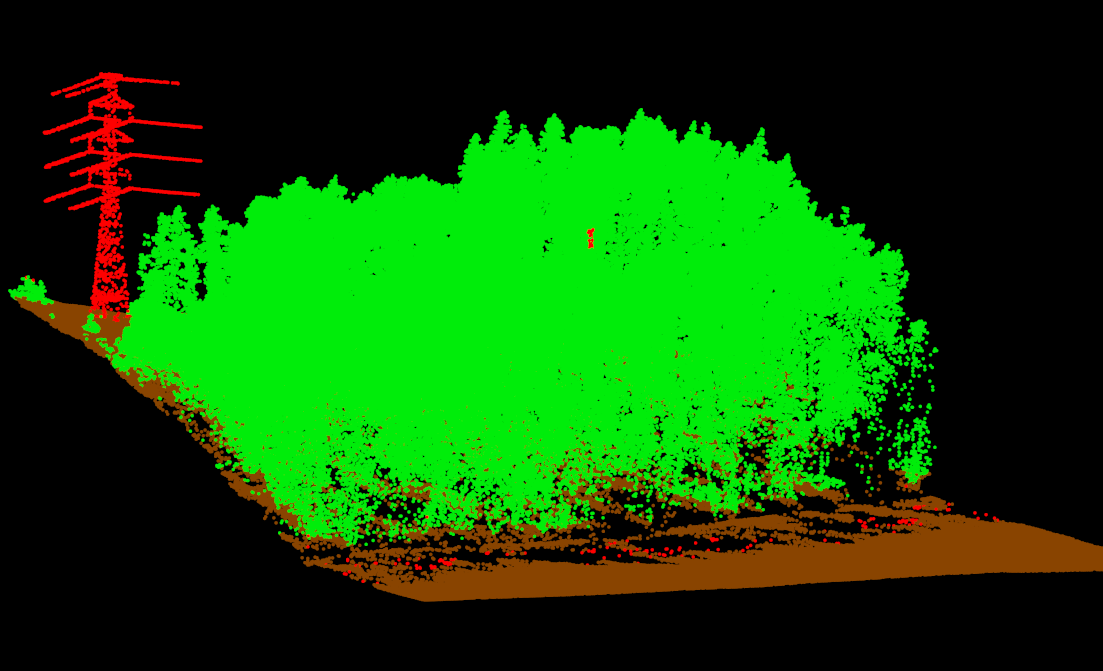}
    }
    \subfloat[h][DOSS\textsuperscript{\dag}]
    {
        \includegraphics[width=0.33\textwidth]{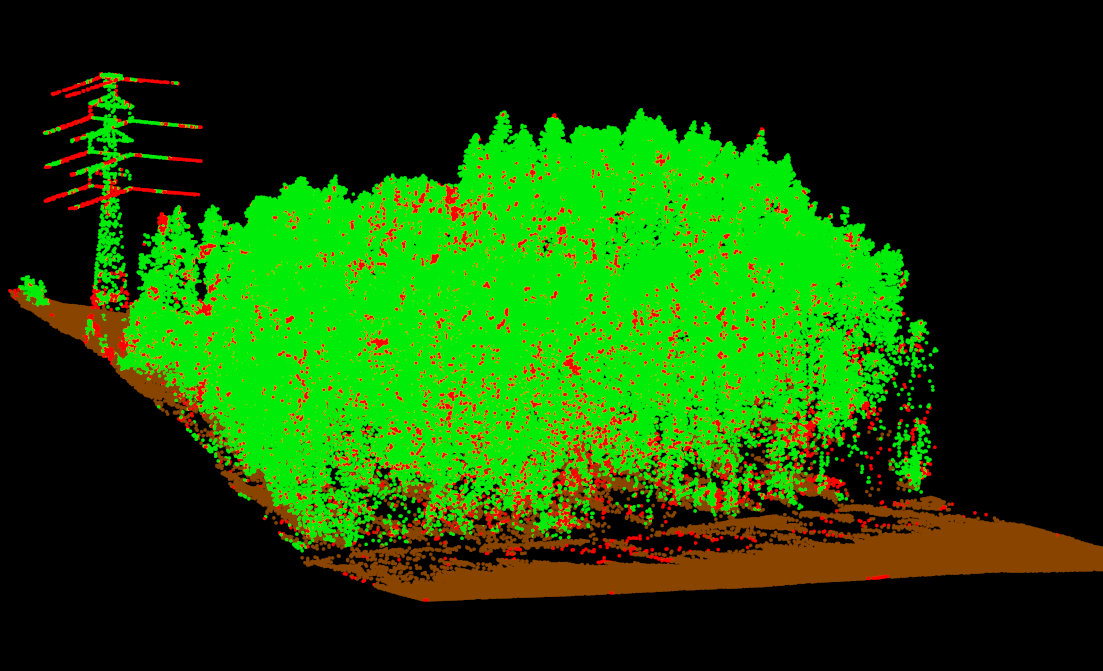}
    }
    \subfloat[i][REAL\textsuperscript{\dag}]
    {
        \includegraphics[width=0.33\textwidth]{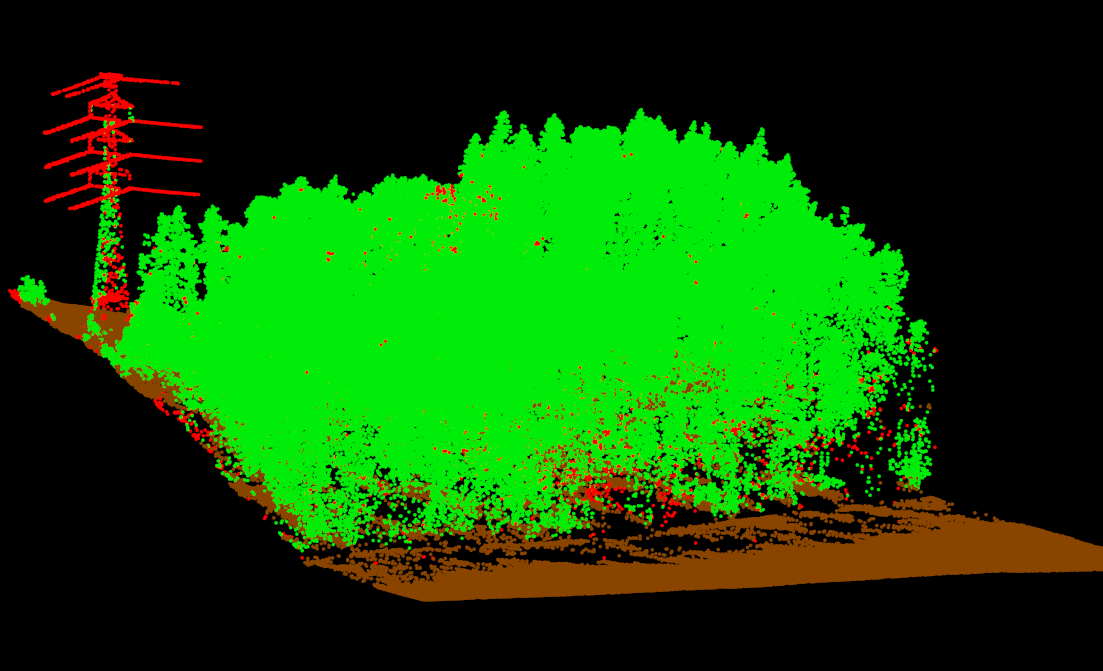}
    }
    \caption{Semantic predictions on ECLAIR dataset. Green is vegetation, brown is ground, and anomalous objects are red. \textsuperscript{\dag} Bottom row is with our reconstruction step, middle is without. The reconstruction step assists our MambaAD network in separating the base of the anomalous object with it's neighbor.}
    \label{fig:EclairADSemantic}    
\end{figure}

In figure \ref{fig:EclairADQualitatitve} we show examples of challenging point clouds where neither method perfectly identifies the anomalous objects.These highlight areas future research can explore. In figure \ref{fig:EclairADSemantic} we show every method on the same scene for direct qualitative comparison. The scene's left side has good RGB values, while on the right they are lacking, demonstrating the importance of method's being able to extract geometric information from the point cloud's co-ordinates. This is especially prevalent in REAL when trained without any synthetic anomalies, where trees on the right side are mislabeled as anomalies.



\newpage

\section{Conclusion}
In this paper we have presented a novel application of reconstruction to large-scale point cloud scenes, as well as the first application of a Mamba backbone to the task of anomaly detection. We demonstrate consistent improvements in anomaly detection, outperforming the comparable methods by over 10\% on AUROC and increasing the AUPR by 80\% on KITTI. On ECLAIR only our Mamba-based anomaly detection architecture was able to achieve an AUPR of 18\% on unknown anomalies, and our reconstruction step then improved it by a further 90\%.

On the secondary task of segmenting known objects, there remains room for improvement, as far as our Mamba-based architecture is concerned. Our inclusion of the reconstruction step for existing methods was a direct approach, simply adding the difference $\Delta\cP$ as extra features for each point. This  method demonstrated improvement for existing methods on the ECLAIR dataset however better incorporation into existing methods could lead to superior performance.

In our ablation tests we explored whether the reconstruction step would do better trained on the scene as a whole, or only on the objects within it. The latter both being more similar to the object defect detection which inspired the method, and also allowing the reconstruction to focus more on the small objects which are more likely to resemble anomalous objects than \eg a section of road. Our results for this specific area are inconclusive, and more research is required.

Finally, while we have shown improvement both in the scalability and performance of anomaly detection, helping the jump to real world applications be less error-prone. The task of identifying truly \textit{unknown} anomalies remains exceedingly difficult, further improvements are still required before LiDAR based systems can reliably handle the chaos of the real world.

\bibliography{bibliography}

\begin{thebibliography}{10}
\providecommand{\url}[1]{\texttt{#1}}
\providecommand{\urlprefix}{URL }
\providecommand{\doi}[1]{https://doi.org/#1}

\bibitem{semanticKitti}
Behley, J., Garbade, M., Milioto, A., Quenzel, J., Behnke, S., Gall, J., Stachniss, C.: {Towards 3D LiDAR-based semantic scene understanding of 3D point cloud sequences: The SemanticKITTI Dataset}. The International Journal on Robotics Research  \textbf{40}(8-9),  959--967 (2021). \doi{10.1177/02783649211006735}

\bibitem{RPCA}
Botterman, H.L., Roussel, J., Morzadec, T., Jabbari, A., Brunel, N.: Robust pca for anomaly detection and data imputation in seasonal time series. In: Nicosia, G., Ojha, V., La~Malfa, E., La~Malfa, G., Pardalos, P., Di~Fatta, G., Giuffrida, G., Umeton, R. (eds.) LOD. pp. 281--295. Springer Nature Switzerland, Cham (2023)

\bibitem{real}
Cen, J., Yun, P., Zhang, S., Cai, J., Luan, D., Tang, M., Liu, M., Yu~Wang, M.: Open-world semantic segmentation for lidar point clouds. In: ECCV. p. 318–334. Springer-Verlag, Berlin, Heidelberg (2022). \doi{10.1007/978-3-031-19839-7_19}

\bibitem{SRMamba}
Chen, C., Ge, W.: Srmamba: Mamba for super-resolution of lidar point clouds (2025)

\bibitem{mamba2}
Dao, T., Gu, A.: Transformers are {SSM}s: Generalized models and efficient algorithms through structured state space duality. In: ICML (2024)

\bibitem{doss}
Deng, W., Chen, X., Yu, Q., He, Y., Xiao, J., Lu, H.: {A Novel Decomposed Feature-Oriented Framework for Open-Set Semantic Segmentation on LiDAR Data}. In: ICRA (2025)

\bibitem{graphSignal}
Dewi, A.O., Ariananda, D.D., Wibowo, S.B.: Graph signal processing for anomaly detection in cooperative power spectrum estimation. In: ICITACEE. pp. 186--191 (2024). \doi{10.1109/ICITACEE62763.2024.10762809}

\bibitem{GANReconstructFromSegmentation}
Di~Biase, G., Blum, H., Siegwart, R., Cadena, C.: Pixel-wise anomaly detection in complex driving scenes. In: CVPR. pp. 16913--16922 (2021). \doi{10.1109/CVPR46437.2021.01664}

\bibitem{mamba}
Gu, A., Dao, T.: Mamba: Linear-time sequence modeling with selective state spaces. arXiv preprint arXiv:2312.00752  (2023)

\bibitem{MambaAD}
He, H., Bai, Y., Zhang, J., He, Q., Chen, H., Gan, Z., Wang, C., Li, X., Tian, G., Xie, L.: Mambaad: Exploring state space models for multi-class unsupervised anomaly detection. In: Globerson, A., Mackey, L., Belgrave, D., Fan, A., Paquet, U., Tomczak, J., Zhang, C. (eds.) NeurIPS. vol.~37, pp. 71162--71187. Curran Associates, Inc. (2024)

\bibitem{VideoMamba}
Li, K., Li, X., Wang, Y., He, Y., Wang, Y., Wang, L., Qiao, Y.: Videomamba: State space model for efficient video understanding (2024)

\bibitem{pointmambaKNNpaper}
Liang, D., Zhou, X., Xu, W., Zhu, X., Zou, Z., Ye, X., Tan, X., Bai, X.: Pointmamba: A simple state space model for point cloud analysis. In: NeurIPS (2024)

\bibitem{pointMamba}
Liu, J., Yu, R., Wang, Y., Zheng, Y., Deng, T., Ye, W., Wang, H.: Point mamba: A novel point cloud backbone based on state space model with octree-based ordering strategy (2024)

\bibitem{3D-UMamba}
Lu, D., Xu, L., Zhou, J., Gao, K., Gong, Z., Zhang, D.: 3d-umamba: 3d u-net with state space model for semantic segmentation of multi-source lidar point clouds. International Journal of Applied Earth Observation and Geoinformation  \textbf{136},  104401 (2025). \doi{https://doi.org/10.1016/j.jag.2025.104401}

\bibitem{eclair2024}
Melekhov, I., Umashankar, A., Kim, H.J., Serkov, V., Argyle, D.: Eclair: A high-fidelity aerial lidar dataset for semantic segmentation. In: CVPR (2024)

\bibitem{CoReSeg}
Nunes, I., Pereira, M.B., Oliveira, H., dos Santos, J.A., Poggi, M.: Conditional reconstruction for open-set semantic segmentation. In: ICIP. pp. 946--950 (2022). \doi{10.1109/ICIP46576.2022.9897407}

\bibitem{ALMRR}
Qu, S., Tao, X., Qu, Z., Gong, X., Zhang, Z., Prasad, M.: Almrr: Anomaly localization mamba on industrial textured surface with feature reconstruction and refinement (2024)

\bibitem{MambaMOS}
Zeng, K., Shi, H., Lin, J., Li, S., Cheng, J., Wang, K., Li, Z., Yang, K.: Mambamos: Lidar-based 3d moving object segmentation with motion-aware state space model (2024)

\bibitem{3DMambaIPF}
Zhou, Q., Yang, W., Fei, B., Xu, J., Zhang, R., Liu, K., Luo, Y., He, Y.: 3dmambaipf: A state space model for iterative point cloud filtering via differentiable rendering (2025)

\bibitem{VisionMamba}
Zhu, L., Liao, B., Zhang, Q., Wang, X., Liu, W., Wang, X.: Vision mamba: Efficient visual representation learning with bidirectional state space model (2024)

\bibitem{cylinder3D}
Zhu, X., Zhou, H., Wang, T., Hong, F., Ma, Y., Li, W., Li, H., Lin, D.: Cylindrical and asymmetrical 3d convolution networks for lidar segmentation. In: CVPR. pp. 9939--9948 (June 2021)

\end{thebibliography}

\end{document}